\documentclass{article}
\usepackage{spconf,amsmath,graphicx,amsfonts, multirow, booktabs,caption,svg,pifont}
\usepackage{hyperref}

\title{Coordinate-based Neural Network for Fourier Phase Retrieval}
\name{Tingyou Li$^{1,*}$, Zixin Xu$^{1,*}$, Yong S. Chu$^2$, Xiaojing Huang$^2$, Jizhou Li$^1$\thanks{$^{*}$ Equal contribution. This work is supported by City University of Hong Kong under grant 9610619.}}
\address{$^1$School of Data Science, City University of Hong Kong, Hong Kong SAR \\
$^2$National Synchrotron Light Source II, Brookhaven National Laboratory, USA
}

\begin{document} 
\ninept
\maketitle
\begin{abstract}

Fourier phase retrieval is essential for high-definition imaging of nanoscale structures across diverse fields, notably coherent diffraction imaging. This study presents the Single impliCit neurAl Network (SCAN), a tool built upon coordinate neural networks meticulously designed for enhanced phase retrieval performance. Remedying the drawbacks of conventional iterative methods which are easiliy trapped into local minimum solutions and sensitive to noise, SCAN adeptly connects object coordinates to their amplitude and phase within a unified network in an unsupervised manner. While many existing methods primarily use Fourier magnitude in their loss function, our approach incorporates both the predicted magnitude and phase, enhancing retrieval accuracy. Comprehensive tests validate SCAN's superiority over traditional and other deep learning models regarding accuracy and noise robustness. We also demonstrate that SCAN excels in the ptychography setting.

\end{abstract}
\vspace{-0.2cm}
\section{Introduction}
\label{sec:intro}
\vspace{-0.1cm}

In imaging systems, detectors typically capture only the amplitude of incoming light, leaving the phase information unrecorded. This omission gives rise to the Fourier phase retrieval challenge, wherein the objective is to reconstruct the lost phase information from the recorded amplitude. Typical applications of Fourier phase retrieval include X-ray crystallography and coherent diffraction imaging (CDI). Successfully retrieving the phase can lead to high-resolution, detailed images of subjects down to the atomic level. In general, Fourier phase retrieval aims to use the measured Fourier magnitude to recover the original object and can be formally written as 
\begin{equation}
\mathcal{F}(k) = |\mathcal{F}(k)|e^{i\psi(k)} = \int^\infty_{-\infty}f(x)e^{-2\pi ikx}dx,
\label{prob}
\end{equation}
where $x$ is an $n$-dimensional spatial coordinate, and $k$ is an $n$-dimensional spatial frequency coordinate. $|F(k)|$ is what we measured on the detector while $\psi(k)$ is lost and needs to be recovered. The $f(x)$ and $|F(k)|$ in the above formulation could have different meanings in different applications. For example, in the transmission mode of CDI, the reconstructed amplitude and phase in $f(x)$ represent the imaginary and real parts of the sample’s refractive index; in the Bragg diffraction mode, its magnitude represents the electron density, and phase denotes the projection of the local deformations of the crystal lattice onto the reciprocal lattice vector Q of the Bragg peak about which the diffraction is measured~\cite{Partial_explain}.

\begin{figure}[!tp]
    \centering
    \setlength{\abovecaptionskip}{0.14cm}
    \includegraphics[width=\linewidth]{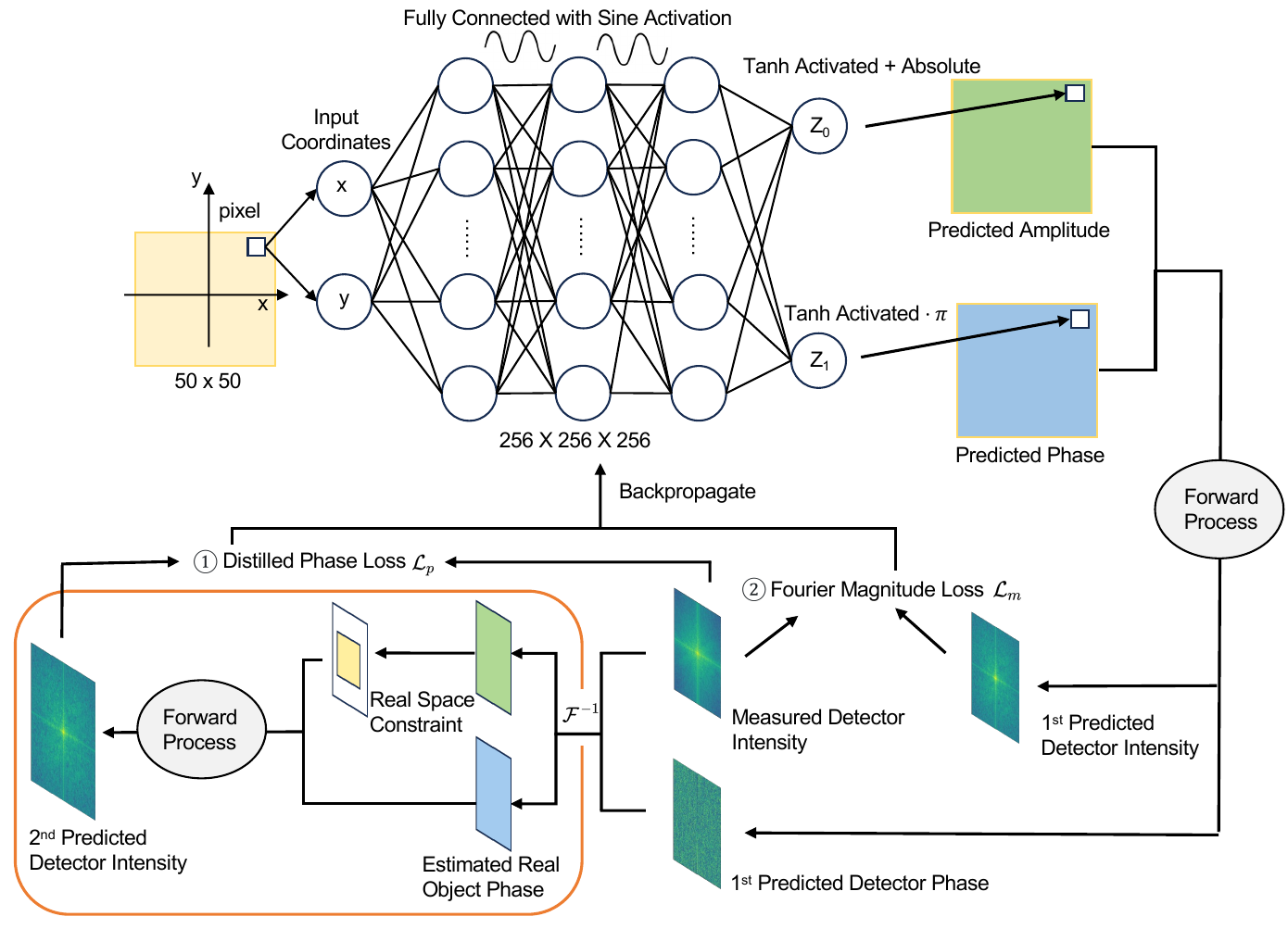} %[width=16cm]
    \caption{Illustration of the proposed SCAN approach for Fourier phase retrieval. }
    \label{fig:Network}
    \vspace{-15pt}
\end{figure}
Despite its critical importance across various domains, the Fourier phase retrieval often remains an elusive challenge due to issues like insufficient oversampling~\cite{oversample,ptycho_net}. Further complications arise from global phase shifts, shift symmetries, and flip symmetries~\cite{end_to_end}, making it an inherently complex non-convex problem. Addressing these challenges, traditional alternating projection techniques like Gerchberg–Saxton (GS)~\cite{GS} and Error Reduction (ER)~\cite{bauschke2002phase} alternate between real and Fourier spaces, enforcing known constraints on their results. However, local optima frequently ensnare these approaches, resulting in sluggish convergence. To mitigate this, methods like HIO~\cite{bauschke2002phase,HIO} have eased these constraints, widening the range of feasible solutions to achieve enhanced performance. Yet, empirical studies indicate that these iterative methods remain susceptible to disruptions from noise~\cite{Deep_phase_cut}. Consequently, they often necessitate extensive iterations and multiple random initializations to obtain reliable results~\cite{bad_tradition}. 

There has been a growing interest in harnessing the capabilities of deep learning to address the limitations inherent in traditional phase retrieval methods. Supervised learning methods, exemplified by approaches like CDI NN~\cite{CDINN}, SiSPRNet~\cite{SiSPRNet} and others~\cite{wu_supervised,3d_supervised}, primarily employ encoder-decoder architectures in CNNs to transform input intensities into complex objects. A notable innovation in this arena is the work presented in \cite{wu_supervised}, which artfully combines supervised learning with transfer learning techniques, refining results in a more adaptable manner. However, even with such advancements, these methods often come with challenges, requiring significant preparatory work and training and potentially faltering when faced with divergent data sets~\cite{Deep_phase_cut}. 

Recognizing the practical requirements, researchers have turned to unsupervised learning methodologies. A prevalent trend within this domain involves utilizing encoder-decoder CNNs, yet without the necessity for pre-training~\cite{ptycho_net,Deep_phase_cut, Autophase,unsupervised_wu,2DIP}. Concurrently, other strategies emerged, leveraging denoising models as a form of regularization~\cite{prDeep}, or introducing specific priors during the recovery process~\cite{phase_prior}. However, the current decoder-encoder-based CNN models may need to tune many parameters before applying them to new data. For example, the convolution stride, the channels, and the block number may need to be changed once the output size is changed. And it could be computationally inefficient once the block increases. Besides, most models~\cite{ptycho_net,Autophase,unsupervised_wu,2DIP} only used Fourier magnitude for supervision during training while the predicted phase is not in use.

%Therefore, one of the main problems with most current deep learning models is that they need to find indirect representations (encoding) for the object so that the neural network can map such representations non-linearly back(decoding) to the object. However, such indirect representations could be inaccurate and time-consuming to find. 

The intrinsic nature of objects, namely their coordinates,  offers an accurate and compact representation. Each coordinate maps directly to its corresponding object amplitude and phase, forming the foundation of the Implicit Neural Representation (INR) concept, a paradigm that has recently been introduced to the research community~\cite{Nerf,SIREN,wavelet}. Prior to our study, this framework demonstrated substantial success in image inversion tasks~\cite{Nerf,Deep_Local_Shapes,NUDF,sun2021coil,added1,added2}. For instance, NeRF~\cite{Nerf} focuses on refining the object's density and color based on multiple viewpoints from various angles. CoIL~\cite{sun2021coil} trains a multilayer perceptron (MLP) to encode the complete measurement field and demonstrates superior performance in sparse-view computed tomography reconstruction. Notably, while DINER~\cite{DINER} and DNF~\cite{DNF} have successfully applied INR for phase recovery, a deeper dive reveals their objective is not Fourier phase retrieval. To our knowledge, there remains an unexplored potential for the INR framework tailored for Fourier phase retrieval. Our contributions can be summarized as follows:

\begin{itemize}
\vspace{-3pt}
    \item We design an unsupervised single coordinate-based neural network (SCAN) based on INR to predict object amplitude and phase simultaneously.
    \vspace{-3pt}
    \item We propose a novel loss function to guide the training phase of SCAN for improved performance.
    \vspace{-3pt}
    \item We benchmark our method against current state-of-the-art (SOTA) algorithms, underscoring its potential through quantitative measurements in CDI and Ptychography experiments.
\end{itemize}

\begin{figure}
\centering
\setlength{\abovecaptionskip}{0.15cm}
\includegraphics[width=\linewidth]{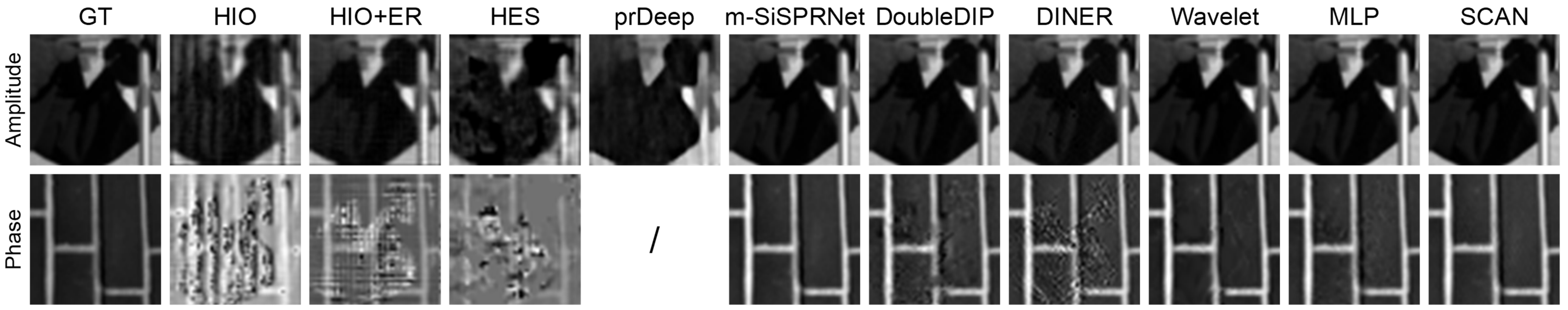}
\caption{Best performance of different methods under noise-free condition}
\label{fig:noisefree}
\vspace{-15pt}
\end{figure}%}

\vspace{-8pt}
\section{Methods}
\label{sec:methods}
\vspace{-2pt}
% \vspace{-0.3cm}
In CDI, the problem can be represented by
\vspace{-2pt}
\begin{equation}
I(\textbf{q})=|\mathcal{F}\{O(\textbf{r})\}|^2, \mathcal{F}\{O(\textbf{r})\}\in\mathbb{C}^{\Tilde{M}\times\Tilde{N}}, O(\textbf{r})\in\mathbb{C}^{M\times N},
\label{prob}
\vspace{-2pt}
\end{equation} where $\Tilde{M}\ge2M-1, \Tilde{N}\ge2N-1$ provide additional information for image recovery. In ptychography, a localized probe is scanned over an extended object, and a series of measurements are collected from overlapping positions:
\vspace{-5pt}
\begin{equation}
I_j(\textbf{q})=|\mathcal{F}\{P(\textbf{r}-\textbf{r}_j)*O(\textbf{r})\}|^2,
\label{prob}
\vspace{-2pt}
\end{equation}
where $\mathcal{F}\{\cdot\}$ is the Fourier transform, $O(\textbf{r})$ represents the object, $P(\textbf{r}-\textbf{r}_j)$ denotes the probe scanning the object at position $j$, $\textbf{r}$ and $\textbf{q}$ are the real space and the Fourier space coordinate. Our idea is to parameterize objects as a (continuous) function represented by $f_\theta(z)$, where $z$ is a set of coordinates, and $f_\theta(\cdot)$ symbolizes a deep neural network with parameters $\theta$. The following equations are formulated in the CDI context, with the ptychography variation conceptualized as scanning CDI employing 
$P(\textbf{r}-\textbf{r}_j)$.

% \subsection{Network Architecture}
\label{ssec:netArch}
The network architecture is illustrated in Fig.~\ref{fig:Network}. The first part is about image coordination, which involves encoding pixels using spatial coordinates. Subsequently, image coordinates are used as inputs for a fully connected network featuring sine activation~\cite{SIREN} to obtain simultaneously the predicted phase and amplitude. The network parameters are iteratively adjusted during training using a meticulously defined loss function; see the details in Sec.~\ref{lossFunc}. 

\vspace{-10pt}
\subsection{Image Coordination}
\vspace{-3pt}
\label{ssec:coordIma}
The entire 2D image is categorized into four quadrants centered around the image's midpoint. Each pixel's horizontal and vertical coordinates are uniformly projected (or normalized) within the range $[-1,1]\times c$, where $c$ is found to be an influential hyperparameter that impacts both the final network outcome and the learning rate. This might be because varying coordinate scales affect the network's initialization output. For smaller values of $c$, the last linear layer output is more homogeneous within a constricted range. Conversely, the output tends to be more normally distributed for larger values. Despite the choice of $c$, the ultimate performance of SCAN remains consistent across different weight initializations. Our experiments found $c$ to be optimally set at 0.1.
% \begin{equation}
% z = z_{normalized} \cdot c
% \label{eqcoordIma}
% \end{equation}$z$ is a set of normalized coordinates.

\begin{figure}
    \centering
    \includegraphics[width=\linewidth]{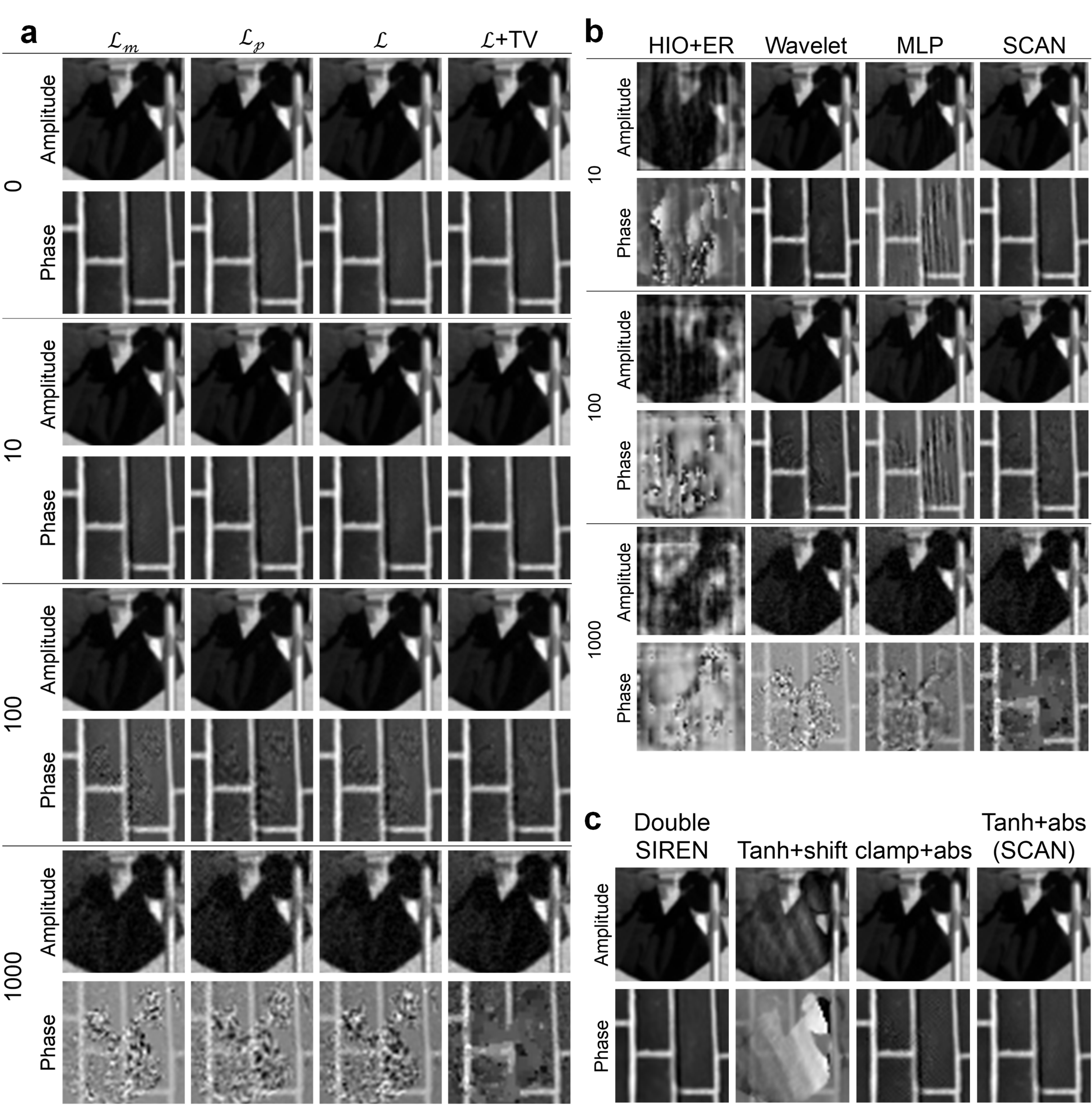}
    \vspace{-15pt}
    \caption{(\textbf{a}) Ablation study of different loss functions under different noise levels ($\sigma=0, 10, 100$, and 1000); (\textbf{b}) comparison of noise robustness across different methods; (\textbf{c}) ablation study of network architectures. }
    \label{fig:ablation}
    \vspace{-16pt}
\end{figure}

\begin{table*}[]
\centering
\setlength{\abovecaptionskip}{0.1cm}
\setlength{\tabcolsep}{6pt} % Default value: 6pt
\renewcommand{\arraystretch}{1}
\resizebox{\linewidth}{!}
{\begin{tabular}{lccccccccccc}
\toprule
        & HIO        & HIO+ER     & HES        & prDeep    & m-SiSPRNet      & Double DIP    & DINER      & Wavelet      & MLP         & SCAN      \\ \midrule
Amp     & 20.12 / 0.54 & 22.52 / 0.85 & 19.30 / 0.49 & 22.49 / 0.65& \textbf{66.81 / 0.9999 }& 45.84 / 0.99    & 35.18 / 0.90 & 53.31 / 0.998  & 46.26 / 0.99  & 55.46 / 0.999 \\
Phase   & 8.12 / 0.17  & 12.94 / 0.32 & 11.74 / 0.26 & -         & 37.46 / 0.997 & 26.95 / 0.83    & 23.08 / 0.68 & 27.84 / 0.92 & 36.43 / 0.94  &\textbf{45.30 / 0.99} 
   \\ \bottomrule
\end{tabular}}
\caption{ The best performance over three random seeds of different methods under noise-free conditions (PSNR / SSIM).}
\label{table:basel}
\end{table*}
\begin{table*}[]
\centering
\vspace{-0.2cm}
\setlength{\abovecaptionskip}{0.1cm}
\resizebox{\linewidth}{!}{
\begin{tabular}{lccccccccccc}
\toprule
          & HIO            & HIO+ER        & HES           & prDeep         & m-SiSPRNet          & Double DIP     & DINER          & Wavelet        & MLP            & SCAN      \\ \hline
Amp       & 13.79$\pm$6.89 & 18.52$\pm$5.01&17.48$\pm$4.68 & 18.50$\pm$4.25 & 29.82$\pm$26.46 &21.43$\pm$17.05 & 24.81$\pm$7.70 & 52.72$\pm$0.75 & 36.88$\pm$17.25&\textbf{63.06$\pm$5.40} \\
Phase     & 9.02$\pm$1.08  & 11.52$\pm$2.67& 10.09$\pm$1.18& -              & 21.10$\pm$11.57 &16.01$\pm$7.74  & 14.72$\pm$5.91 & 25.13$\pm$1.95 & 23.90$\pm$10.86&\textbf{42.43$\pm$2.43}    \\ \bottomrule
\end{tabular}}
\caption{ Stability of different methods in mean$\pm$std (PSNR).}
\vspace{-15pt}
\label{table:baselstd}
\end{table*}

\vspace{-6pt}
\subsection{Backbone of Main Network}
\vspace{-3pt}
\label{Backbone} 
The fully connected network consists of three hidden layers, each with 256 units, and a concluding output layer with two units. While hidden layers are integrated with sine activation adapted from SIREN~\cite{SIREN}, the final output layer is followed by a Tanh activation function in our approach. The primary output unit predicts the object's amplitude, whereas the second determines its phase. Therefore, an absolute operation is performed after activating the first output unit. The second output is multiplied by $\pi$ to ensure the phase range is scaled to $[-\pi, \pi]$. The process can be written as:
\begin{equation}
\widehat{O(\textbf{r})} = \left|f_{\theta|1}(z)\right|e^{if_{\theta|2}(z)\pi},
\label{eqpara}
\vspace{-5pt}
\end{equation}
where $z$ is coordinate normalization from $\textbf{r}$ and $f_{\theta|1}(z)$, $f_{\theta|2}(z)$ represent the outputs from the first and second channel, respectively.

It's worth highlighting that SCAN utilizes just one fully connected network. This design choice means SCAN has a limited set of hyperparameters — specifically, the learning rate, layer size, and number of layers. In contrast, approaches with separate networks for amplitude and phase outputs often demand more intricate tuning, such as individualized learning rates and layer adjustments. As a result, SCAN benefits from having fewer parameters, streamlining the optimization process.

\vspace{-10pt}
\subsection{Loss Function}
\vspace{-5pt}
\label{lossFunc}

\textbf{2.3.1 Fourier Magnitude Loss ($\mathbf{\mathcal{L}_m}$)}. Upon forward propagation through the network, the predicted amplitude and phase undergo additional forwarding to derive the predicted Fourier magnitude and phase. The primary loss metric evaluates the discrepancy between the predicted Fourier magnitude and the actual measurements. This form of loss assessment is commonly employed in unsupervised end-to-end methodologies~\cite{ptycho_net,Autophase,unsupervised_wu,2DIP} because it directly quantifies the accuracy of the Fourier magnitude prediction. We write it as:
\begin{equation}
\mathcal{L}_m = \left\||\mathcal{F}\{\widehat{O(\textbf{r})}\}|-\sqrt{I(\textbf{q})}\right\|_2^2. 
\label{eqintenloss}
% \vspace{-5pt}
\end{equation}

\hspace{-15pt}\textbf{2.3.2 Distilled Fourier Phase Loss ($\mathbf{\mathcal{L}_p}$)}. We further leverage the predicted phase by incorporating it into the loss calculation after suitable adjustments. Inspired by ER~\cite{HIO}, which employs iterative processes between real and Fourier spaces, we integrate the predicted Fourier phase $u$ with measured detector intensity $I$ to transition back to the real space, as illustrated in Fig.~\ref{fig:Network}. Subsequently, we apply recognized constraints in the real space, denoted by $\odot S$, to this transformed object. These constraints enforce that the padding area should be zero and the object amplitude must reside within the 0-1 range. After imposing these constraints on the transformed object, we transition it forward to the Fourier space, deriving the Fourier magnitude for the second time. The secondary loss evaluates the disparity between this second-time predicted Fourier magnitude and the actual measurement.

The underlying rationale for the second loss $\mathcal{L}_p$ is that if the predicted Fourier phase is accurate, the merging of this phase with the true Fourier magnitude, coupled with the back-and-forth transitions between the Fourier and real spaces under established constraints, should yield a Fourier magnitude consistent with the measured one. The equations are detailed below. In this context, $\widehat{O_a}$ and $\widehat{O_p}$ represent the amplitude and the phase of the predicted object, respectively. They arise from the fusion of the measured detector intensity $I$ and the predicted detector phase $u$.

% The intuition for the second loss $\mathcal{L}_p$ is that assumed the predicted Fourier phase is correct, then after combining with actual Fourier magnitude and iterating between Fourier space and real space with known constraint, it should still remain the same as measured Fourier magnitude. Equations are shown below. Here $\widehat{O_a}$ and $\widehat{O_p}$ respectively represent the amplitude and the phase of the predicted object that derived from the combination of measured magnitude $I$ and predicted detector phase $u$ separately.

\vspace{-5pt}
\begin{equation}
u = angle(\mathcal{F}\{\widehat{O(\textbf{r})}\}),
\label{eqprephase}
\end{equation}
\begin{equation}
\widehat{O_a} = |\mathcal{F}^{-1}\{\sqrt{I}e^{iu}\}|\odot S,\  \widehat{O_p} = angle(\mathcal{F}^{-1}\{\sqrt{I}e^{iu}\}),
\label{eqpreamp}
\end{equation}
\begin{equation}
\mathcal{L}_p = \left\||\mathcal{F}\{\widehat{O_a}e^{i\widehat{O_p}}\}|-\sqrt{I}\right\|_2^2. 
\label{eqprede}
\end{equation}
\textbf{2.3.3 Final Loss Function}. Our ultimate training loss constitutes a blend of these individual losses. Both losses $\mathcal{L}_m$ and $\mathcal{L}_p$ are deemed equally significant during training. However, given that their magnitudes differ, it's necessary to preface each with an appropriate weight. The cumulative loss is expressed as $\mathcal{L} = W_1\times \mathcal{L}_m + W_2\times \mathcal{L}_p$. In subsequent experiments, considering the initial ratio of Fourier magnitude loss to Distilled Fourier phase loss to be approximately 7:3, we set $W_1=0.3$ and $W_2=0.7$ for simplicity. Moreover, a Total Variation (TV) loss on the predicted phase is further incorporated for scenarios with noise disruption. By incorporating the TV loss, we promote smoother phase transitions and improve the overall quality of the phase predictions.

\vspace{-0.25cm}
\section{Experiments}
\label{sec:Experiments}
\vspace{-0.2cm}
% 3. 3. EXPERIMENTS AND RESULTS
\subsection{Experimental setup}
\vspace{-5pt}
\label{Dataset}

Two commonly employed test images are utilized in our subsequent experiments: the \textit{cameraman} and the \textit{brick}. These images are resized to dimensions $256\times256$ and are normalized using min-max normalization to the range [0,1] prior to experimentation. We will evaluate the performance of the proposed SCAN method in comparison with state-of-the-art techniques in both the CDI and Ptychography settings. Analysis of our approach, including ablation studies, will also be conducted.

\hspace{-15pt}\textbf{Coherent Diffraction Imaging (CDI)}. 
In the existing studies, the oversample rate for most experiments is typically set around 2, with the actual object positioned ambiguously within the designated area. Consequently, a shifted symmetry may be observed during the recovery phase, as noted in~\cite{end_to_end}. Moreover, areas outside the object could potentially register values due to the quality of recovery. This scenario complicates the adoption of quantitative quality metrics like PSNR or SSIM. While the shrinkwrap method~\cite{Shrinkwrap}, which reduces minor values to zero, might provide a solution, the threshold for shrinkwrap can significantly vary across experiments, rendering comparisons unfair. Hence, we set the known object size to match the unpadded area in subsequent experiments to circumvent the shift issue and concentrate solely on the model's recovery capability. We chose an oversample rate of approximately 5 for all benchmarked methods. Our simulated 2D object amplitude is a segment of the cameraman image ($50\times50$ pixels), and the object phase derives from a portion of the brick image of the same size (Fig.~\ref{fig:noisefree}).

In our experimentation, we will compare traditional alternating projection techniques such as HIO, HIO+ER~\cite{HIO}, and HES~\cite{Shrinkwrap} as well as deep models like SiSPRNet~\cite{SiSPRNet}, prDeep~\cite{prDeep}, and DINER~\cite{DINER}. Given that our model employs sine activation function, we also assessed other potential backbones for this task. The Deep Image Prior~\cite{ulyanov2018deep}, MLP with a positional encoder, and a recent wavelet approach~\cite{wavelet} are also examined. It's worth mentioning that SiSPRNet~\cite{SiSPRNet} operates as a supervised model. For the purposes of this study, we only adapted its architectural structure (CNN-based decoder-encoder) for comparison, referring to it as m-SiSPRNet. Additionally, only the real recovery component in prDeep~\cite{prDeep} is factored into our comparative analysis.

With the above settings, we are interested in the following aspects: (1) Recovery performance under noise-free conditions; (2) Stability of the models; (3) Noise robustness; (4) Ablation studies of our approach, such as the loss function, network architectures etc. 

% \begin{itemize}
%     \item Noise-free condition: No noise in the simulated square root of intensity.
%     \item Stability of the models: Models are run with the same parameters on three different seeds
%     \item Gaussian noise condition: Three levels (10/255, 100/255, 1000/255) of noise are added to simulated square root of intensity
%     \item Ablation study of the loss function
%     \item Ablation study of different model structures (Tanh, Double SIREN) 
% \end{itemize}

% \begin{figure}
%     \centering
%     \includegraphics[width=5cm]{comnoise.png}%[width=10cm]
%     \caption{Performance of Models under Noise Condition}
%     \label{fig:noise}
% \end{figure}

\hspace{-15pt}\textbf{Ptychography}. Distinct from the CDI setting, the ptychography experiment aims to recover the whole image ($256\times 256$ pixels) with multiple measured intensities. It has a simulated probe with a dimension of $28\times 28$ derived from ~\cite{revealing} to scan different positions of the whole image. Since most existing networks are tailored for CDI, our model's benchmark will be ePIE. Notably, the overlap rate plays a pivotal role in determining the quality of reconstruction and the computational time. Hence, we evaluate the model's performance across different overlap rates, specifically at 30\%, 50\%, and 70\%.

\vspace{-10pt}
\subsection{Model Training}
\vspace{-3pt}
\label{ExperimentalSetup}
We evaluated SCAN on a server with one NVIDIA GeForce RTX 3090 GPU. All models use Adam optimizer for training. Only the learning rate is changed in optimizer hyperparameters.  For SCAN, the learning rate is fixed to be $0.8\times 10^{-4}$. For other methods, learning rates are adjusted to obtain the best performance. The batch size is set to be 1 as it inputs the whole image for each epoch.  

% 不同noise 不同方法 SCAN有修改，还没放上来 放了
\begingroup
\setlength{\tabcolsep}{2.2pt} % Default value: 6pt
\renewcommand{\arraystretch}{1.15} % Default value: 1
\begin{table}
\footnotesize
\centering
\vspace{-0.4cm}
\setlength{\abovecaptionskip}{0.1cm}
% \resizebox{\columnwidth}{!}{
\begin{tabular}{@{}lccccc@{}}
\toprule
       &                     & HIO+ER      & Wavelet     & MLP        & SCAN       \\ \hline
\multirow{2}{*}{10}   & Amp & 16.14 / 0.36 & 52.83 / 0.998 & 32.99 / 0.71   & \textbf{61.13 / 0.999} \\
                      & Phase &11.22 / 0.12 & 26.74 / 0.92  & 18.25 / 0.57  & \textbf{35.42 / 0.98}  \\\hline
\multirow{2}{*}{100}  & Amp & 14.51 / 0.21& 42.62 / 0.98  & 31.30 / 0.70    & \textbf{42.93 / 0.98 } \\
                      & Phase & 9.21 / 0.09  & \textbf{26.48} / 0.84  & 18.81 / 0.55 & 26.42 / \textbf{0.85} \\\hline
\multirow{2}{*}{1000} & Amp & 11.65 / 0.22 & 29.03 / 0.75  & 23.81 / 0.58   & \textbf{29.19 / 0.76}\\
               & Phase      & 8.22 / 0.09  & 9.74 / 0.25   & 10.61 / 0.28   & \textbf{17.18 / 0.58 } \\ \bottomrule
\end{tabular}%}
\caption{Performance of phase retrieval methods under noisy conditions (PSNR / SSIM)}
\label{table:baselnoise}
\vspace{-0.5cm}
\end{table}
\endgroup

\vspace{-0.2cm}
\section{Results}
\label{sec:Results}
% \vspace{-0.2cm}
\subsection{Results of CDI setting}
\vspace{-3pt}
\label{CDI}
\textbf{Noise Free Condition}. Under noise-free conditions, Tab.~\ref{table:basel} and Fig.~\ref{fig:noisefree} show the best results for each method out of 3 random realizations (seeds). m-SiSPRNet achieved the best performance in amplitude recovery, followed by SCAN and wavelet. For object phase recovery, SCAN performed the best, then MLP and m-SiSPRNet. Phase is found to be harder than amplitude to recover. The same model could have quite different recovery abilities on these two properties. For instance, m-SiSPRNet's phase recovery is 30 dB lower than its amplitude result. However, SCAN balanced the results and achieved good performance on both.
% Under the noise-free condition, the best performance within three random experiments of different models is reported in Tab.~\ref{table:basel} and Fig.~\ref{fig:noisefree}. Although SiSPRNet performs better in amplitude, its mean is roughly 30 dB lower than ours with much higher instability. In the rest of the models, our model performs best following with Wavelet and MLP, and phase is nearly 9 dB higher than MLP, the second best.

\hspace{-15pt}\textbf{Stability}.
Tab.~\ref{table:baselstd} shows the mean and standard deviation of 3 seeds' experiment results under the noise-free condition. SCAN has the best amplitude and phase recovery performance when considering all seeds. It is noticed that decoder-encoder-based models, m-SiSPRNet and Double DIP, have a high variance in results. SCAN had a relatively lower variance, demonstrating its high stability.

\hspace{-15pt}\textbf{Noise Robustness}. Tab.~\ref{table:baselnoise} and Fig. 3(a) show the restoration performance under different noise levels. In most cases, SCAN outperforms all methods, especially in retrieving the object phase. 

\hspace{-15pt}\textbf{Ablation Study of Loss Function}.
As shown in Tab.~\ref{table:loss} and Fig. 3(b), the $\mathcal{L}_p$ recovers the phase better than $\mathcal{L}_m$ in most cases. The model with the combination of these two losses improves the recovery quality of the phase compared with the one with $\mathcal{L}_m$, and the amplitude recovery outperforms both. The performance is further enhanced after adding a TV loss, mainly in the phase part. 

\begingroup
\setlength{\tabcolsep}{2pt} % Default value: 6pt
\renewcommand{\arraystretch}{1.15}
\begin{table}[!th]
\footnotesize
\vspace{-0.3cm}
\setlength{\abovecaptionskip}{0.06cm}
\centering
% \resizebox{\columnwidth}{!}{
\begin{tabular}{lccccc}
\toprule
                        &           & \begin{tabular}[c]{@{}l@{}}$\mathcal{L}_m$\end{tabular} & \begin{tabular}[c]{@{}l@{}}$\mathcal{L}_p$\end{tabular} & $\mathcal{L}$    & \begin{tabular}[c]{@{}l@{}}$\mathcal{L}$+TV\end{tabular} \\ \hline
\multirow{2}{*}{0}    & Amp & 50.50 / 0.99                                               & 52.23 / 0.99                                                     & 63.06 / 0.999 & \textbf{63.06 / 0.999}                                            \\
                        & Phase     & 32.88 / 0.93                                               & 36.63 / 0.94                                                     & 42.43 / 0.99  & \textbf{42.43 / 0.999}                                             \\\hline
\multirow{2}{*}{10}   & Amp & 53.97 / 0.996                                              & 52.81 / 0.998                                                    & 59.78 / 0.999 & \textbf{61.13 / 0.999}                                            \\
                        & Phase     & \textbf{38.46} / 0.96                                               & 33.41 / 0.93                                                     & 34.72 / 0.98  & 35.42 / \textbf{0.98}                                             \\\hline
\multirow{2}{*}{100}  & Amp & 42.33 / 0.97                                               & 42.71 / 0.97                                                     & 42.58 / 0.97  & \textbf{42.93 / 0.98}                                             \\
                        & Phase     & 19.01 / 0.74                                               & 23.60 / 0.80                                                     & 21.48 / 0.79  & \textbf{26.42 / 0.85}                                             \\\hline
\multirow{2}{*}{1000} & Amp & 28.53 / 0.74                                               & 26.61 / 0.67                                                     & 28.59 / 0.75  & \textbf{29.19 / 0.76}                                             \\
                        & Phase     & 10.35 / 0.27                                               & 10.18 / 0.30                                                     & 11.10 / 0.32  & \textbf{17.18 / 0.58}                                             \\ \bottomrule
\end{tabular}%}
\caption{ Ablation study of different loss function. (PSNR / SSIM)}
\label{table:loss}
% \vspace{-0.3cm}
\end{table}
\endgroup

\begin{table}[!th]
% \vspace{-0.4cm}
\centering
\setlength{\abovecaptionskip}{0.06cm}
\resizebox{\columnwidth}{!}{
\begin{tabular}{lcccc}
\toprule
          & Double SIREN  & Tanh+shift  & clamp+abs & Tanh+abs (SCAN) \\ \midrule
Amp & 43.75 / 0.88  & 12.77 / 0.28  & 62.44 / 0.999  & \textbf{63.06 / 0.999}\\
Phase     & 25.27 / 0.73 & 9.53 / 0.22 & 37.01 / 0.986 & \textbf{42.43 / 0.99 }\\ 
Time(s)   & 448.02 &-   &-                      &\textbf{324.52}   \\\bottomrule
\end{tabular}}
\caption{ Ablation study of different model structures. (PSNR / SSIM)}
\label{table:structure}
\vspace{0.4cm}
\end{table}

\begingroup
\setlength{\tabcolsep}{1.2pt} % Default value: 6pt
\renewcommand{\arraystretch}{1} % Default value: 1
\begin{table}[h!]
\vspace{-0.5cm}
\resizebox{\columnwidth}{!}{
\setlength{\abovecaptionskip}{0.06cm}
\centering
\begin{tabular}{@{}lccccccccccc@{}}
\toprule
           &\multicolumn{3}{c}{30\%} & &\multicolumn{3}{c}{50\%}  & &\multicolumn{3}{c}{70\%}      \\ \hline
           & Amp & Phase & Time(s)  & & Amp & Phase & Time(s)   & & Amp & Phase & Time(s)  \\ \cmidrule(lr){2-4} \cmidrule(lr){6-8} \cmidrule(l){10-12}
ePIE       &9.75 / 0.20&11.03 / 0.20&76.58& &10.08 / 0.17&11.14 / 0.23&141.10& &29.70 / 0.90&17.72 / 0.80&394.69\\ 
SCAN       &\textbf{23.06 / 0.82}&\textbf{15.38 / 0.70}&\textbf{24.14}& &\textbf{25.15 / 0.81}&\textbf{16.02 / 0.71}&\textbf{35.77}& &\textbf{34.65 / 0.98}&\textbf{21.23 / 0.87}&\textbf{237.08}\\ \bottomrule
\end{tabular}}
\vspace{-0.2cm}
\caption{ Ptychography reconstruction of ePIE and SCAN when the overlap rate is 30\%, 50\%, and 70\%. (PSNR / SSIM)}
\label{table:ptychtest}
\vspace{-0.2cm}
\end{table}
\endgroup

% \begin{table}[]
% \begin{tabular}{lccc|ccc|ccc}
% \toprule
%      & \multicolumn{3}{c|}{30\%} & \multicolumn{3}{c|}{50\%} & \multicolumn{3}{c}{70\%} \\ \hline
%      & Amp   & Phase  & Time(s)  & Amp   & Phase  & Time(s)  & Amp  & Phase  & Time(s)  \\ \cline{2-10} 
% ePIE & 10.31/0.21&11.04/0.25&161.18&13.12/0.35&11.49/0.35&315.44&35.26/0.98&12.15/0.56&724.08\\ 
% SCAN &\textbf{25.17/0.82}&\textbf{19.40/0.77}&\textbf{45.21}&\textbf{27.94/0.93}&\textbf{24.59/0.89}&\textbf{72.24}&\textbf{36.69/0.998}&\textbf{35.65/0.98}&\textbf{341.76}\\ \bottomrule
% \end{tabular}
% \end{table}
\hspace{-15pt}\textbf{Ablation Study of Different Model Structures}.
Tab.~\ref{table:structure} and Fig.~\ref{fig:ablation} present several potential model implementations. The terms ``clamp" and ``shift'' are utilized to denote alternative approaches to the Tanh activation function and absolute operation, respectively. Specifically, ``clamp" refers to constraining the value within the range of $[-1, 1]$, while ``shift" involves scaling the value $x$ to the range of $[0, 1]$ using the transformation $(x + 1)/2$. Additionally, we explored utilizing two separate networks, one for object amplitude and the other for phase representation. The results demonstrate that our current implementation in SCAN outperforms all other approaches.

\begin{figure}[!t]
    \centering
    \setlength{\abovecaptionskip}{0.15cm}
    % \vspace{-0.6cm}
    % \rotatebox{-90}{
    \includegraphics[width=8cm]{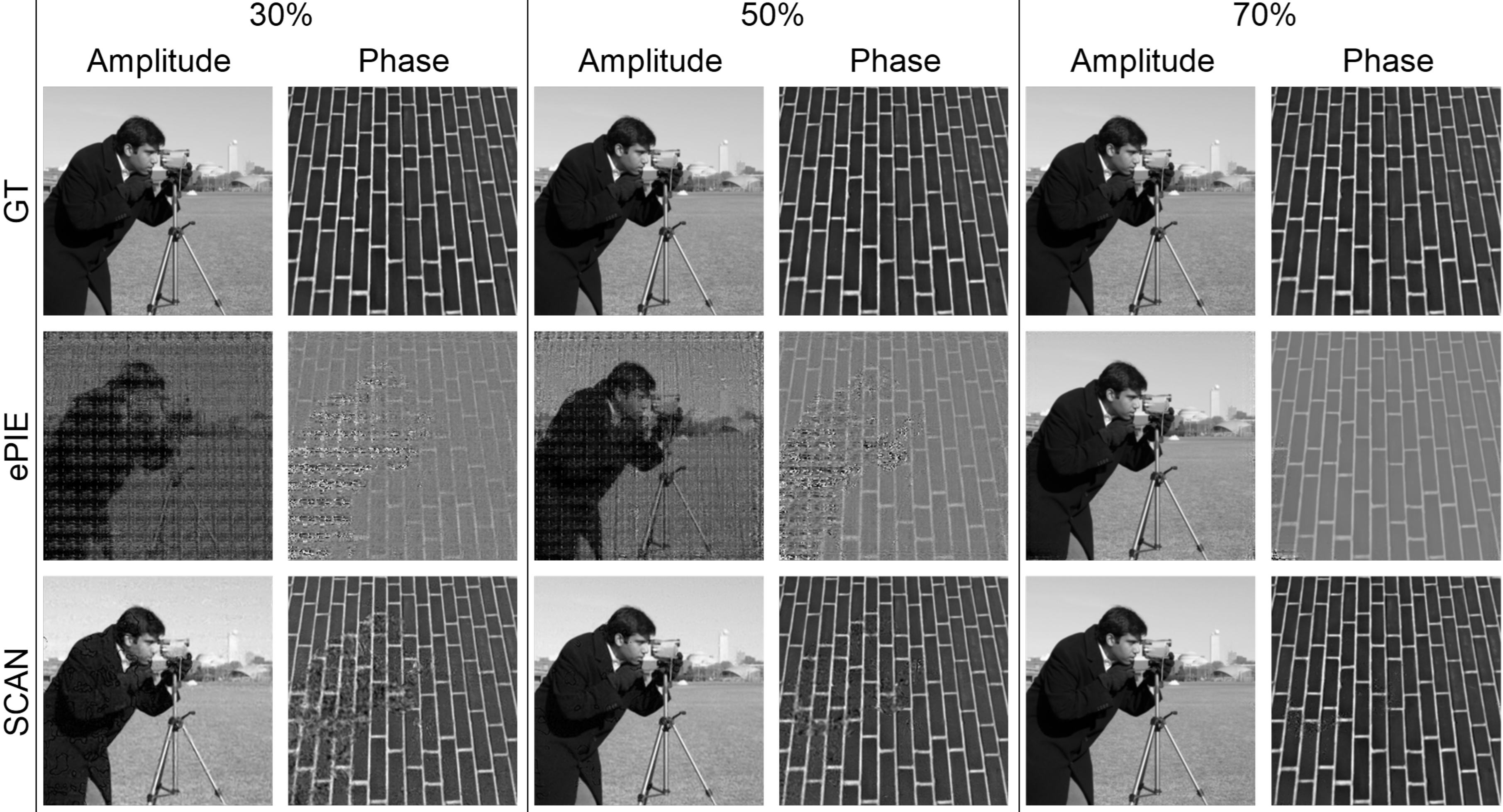}
    \caption{Ptychographic reconstruction comparisons between ePIE and SCAN at overlap rates of 30\%, 50\%, and 70\%.}
    \label{fig:ptyc}
    \vspace{-0.5cm}
\end{figure}

\vspace{-0.2cm}
\subsection{Results of Ptychography setting}
\label{Ptychography}
As shown in Tab.~\ref{table:ptychtest} and Fig.~\ref{fig:ptyc}, SCAN exhibits superior quality in reconstructions at low overlap rates of 30\% and 50\%. In contrast, ePIE necessitates an overlap in the range of 70\% to achieve commendable results, as referenced in \cite{ePIE}. Notably, the reconstruction of SCAN is substantially faster than ePIE. Additionally, the SCAN method appears adept at mitigating periodic artifacts, even under conditions of minimal overlap.

% SCAN recovers at a high quality when the rates are 30\% and 50\%, while ePIE with a good result requires an overlap of 60\%-70\%~\cite{ePIE}. And the reconstruction speed is much higher than the classical one. The SCAN method seems to be able to suppress the periodic artefact, even in low overlaping condition.
\vspace{-0.3cm}
\section{Conclusions}
\label{sec:conclu}
% \vspace{-0.2cm}

We have introduced SCAN, a coordinate-based neural network framework devised to advance the precision and efficiency of Fourier phase retrieval problems. Through comprehensive experimental results and ablation studies, SCAN demonstrated its robustness, especially in challenging noisy scenarios, outperforming existing methodologies. The unified network architecture and newly introduced loss function provide distinct advantages over traditionally employed unsupervised models in the domain. Extending SCAN to address broader phase retrieval challenges and applications in 3D Bragg Ptychography remains a promising avenue for future exploration.

\clearpage

% \bibliographystyle{IEEEbib}
% \bibliography{Z_Main_Article}

\end{document}